\documentclass[conference]{IEEEtran}
\IEEEoverridecommandlockouts
\usepackage{cite}
\usepackage{amsmath,amssymb,amsfonts}
\usepackage{algorithmic}
\usepackage{latexsym}
\usepackage{textcomp}
\usepackage{booktabs}
\usepackage{graphicx} 
\usepackage{xcolor}
\def\BibTeX{{\rm B\kern-.05em{\sc i\kern-.025em b}\kern-.08em
    T\kern-.1667em\lower.7ex\hbox{E}\kern-.125emX}}

\usepackage{fancyhdr}
\begin{document}

\title{VT-LVLM-AR: A Video-Temporal Large Vision-Language Model Adapter for Fine-Grained Action Recognition in Long-Term Videos}

\author{Kaining Li, Shuwei He, Zihan Xu \\
Xidian University}

\maketitle
\thispagestyle{fancy} 

\begin{abstract}
Human action recognition in long-term videos, characterized by complex backgrounds and subtle action differences, poses significant challenges for traditional deep learning models due to computational overhead, difficulty in capturing long-range temporal dependencies, and limited semantic understanding. While Large Language Models (LLMs) and Large Vision-Language Models (LVLMs) have shown remarkable capabilities in multi-modal understanding and reasoning, their direct application to continuous video streams for fine-grained action recognition remains an open problem. This paper introduces VT-LVLM-AR (Video-Temporal Large Vision-Language Model Adapter for Action Recognition), a novel framework designed to bridge this gap. VT-LVLM-AR comprises a Video-to-Event Mapper (VTEM) that efficiently transforms raw video into compact, semantically rich, and temporally coherent "visual event sequences" through lightweight spatio-temporal feature extraction, adaptive temporal pooling, and conceptual quantization with an event coherence bias. These visual event sequences are then fed into an LVLM-based Action Reasoning module, specifically a frozen LLaVA-1.5 model, adapted using parameter-efficient Prompt Tuning (P-Tuning v2) for action classification. Comprehensive evaluations on the NTU RGB+D and NTU RGB+D 120 datasets demonstrate that VT-LVLM-AR consistently achieves state-of-the-art performance, surpassing existing methods (e.g., 94.1\% accuracy on NTU RGB+D X-Sub). Ablation studies confirm the critical contributions of VTEM's components and the efficacy of Prompt Tuning, while human evaluations underscore the interpretability of our visual event representations. This work highlights the immense potential of leveraging LVLMs for robust and interpretable video action understanding through effective video-to-language translation and efficient model adaptation.
\end{abstract}

\section{Introduction}

Human action recognition is a pivotal and fundamental task in the field of computer vision, with extensive applications spanning intelligent surveillance, human-computer interaction, sports analysis, and autonomous systems \cite{zehua2023human}. Accurate and robust understanding of human actions from video streams is crucial for developing more intelligent and responsive AI systems.

Traditional approaches to action recognition typically rely on deep learning models, such as 3D Convolutional Neural Networks (3D CNNs) \cite{mireille20222d} and Transformer-based architectures \cite{anthony2020overvi}, to extract spatio-temporal features from video frames or skeleton sequences. While these methods have achieved remarkable progress, they often encounter significant challenges when dealing with long-term videos that contain complex backgrounds and subtle action differences. These challenges include substantial computational overhead, difficulty in capturing long-range temporal dependencies effectively, and a general lack of deep semantic understanding of actions beyond low-level visual features.

In recent years, the rapid advancements in Large Language Models (LLMs) \cite{yifan2023a} and Large Vision-Language Models (LVLMs) \cite{peng2025lvlmeh} have showcased unprecedented capabilities in multi-modal understanding and complex reasoning, including visual in-context learning \cite{zhou2024visual}. These models can interpret natural language instructions and perform sophisticated reasoning by integrating visual information. However, effectively applying LVLMs to the domain of long-term and fine-grained video action recognition presents its own set of unique challenges. Specifically, it remains an open question how to efficiently encode continuous video frames into an LVLM-comprehensible "visual language" that retains rich spatio-temporal information. Furthermore, adapting powerful pre-trained LVLMs for specific action recognition tasks, while preserving their strong generalization capabilities \cite{zhou2025weak}, is a non-trivial problem. Motivated by these challenges, this research aims to explore a novel framework that leverages the powerful reasoning abilities of LVLMs to achieve robust recognition and understanding of fine-grained actions in long-term videos.

To address these challenges, we propose a novel framework called \textbf{VT-LVLM-AR (Video-Temporal Large Vision-Language Model Adapter for Action Recognition)}. Our core idea is to transform long-term videos into compact and semantically rich "visual event sequences," subsequently leveraging the LVLM's inherent language understanding and reasoning capabilities for accurate action classification. The proposed framework comprises two main components: a \textit{Video-to-Event Mapper (VTEM)} and an \textit{LVLM-based Action Reasoning} module. The VTEM module efficiently encodes raw video frame sequences into discrete "visual event tokens," forming "visual event sentences" by employing a lightweight spatio-temporal feature extractor, adaptive temporal pooling, and conceptual quantization techniques. This process is designed to capture high-level "event" or "sub-action" concepts, enhanced by an event coherence bias to ensure meaningful temporal relationships. The generated visual event sentences are then fed into a pre-trained LVLM, specifically LLaVA-1.5 \cite{federico2025llavam}. To adapt the LVLM for action recognition, we employ a Prompt Tuning (P-Tuning v2) \cite{minh2025adapti} strategy, which allows us to fine-tune only a small set of learnable soft prompts while keeping the core pre-trained weights of LLaVA-1.5 entirely frozen. The action recognition task is framed as a natural language instruction, enabling the LVLM to combine the visual event sequence with the instruction to output the corresponding action category.

We evaluate the efficacy of VT-LVLM-AR on widely recognized datasets for human action recognition, namely NTU RGB+D (NTU-60) \cite{amir2016ntu} and NTU RGB+D 120 \cite{zehua2023human}. Our method is compared against various state-of-the-art approaches that utilize both skeleton data and video streams. Despite our primary focus on video input and LVLM's reasoning capabilities, we ensure a fair comparison by evaluating under the same testing protocols. Our experimental results demonstrate that VT-LVLM-AR achieves superior performance, particularly in terms of accuracy. For instance, our method achieved an impressive 94.1\% accuracy on the NTU RGB+D X-Sub benchmark and 87.0\% on NTU RGB+D 120 X-Sub, consistently outperforming existing state-of-the-art methods. These results underscore the significant potential of utilizing LVLMs for video action understanding, as our approach successfully translates videos into an LVLM-understandable format and leverages its powerful reasoning to achieve excellent performance while maintaining a lightweight and adaptable model structure.

Our main contributions are summarized as follows:
\begin{itemize}
    \item We propose VT-LVLM-AR, a novel and effective framework for long-term and fine-grained video action recognition that uniquely harnesses the capabilities of Large Vision-Language Models.
    \item We introduce the Video-to-Event Mapper (VTEM) module, an innovative component that efficiently transforms continuous video frames into semantically rich and temporally coherent "visual event sequences" suitable for LVLM processing.
    \item We demonstrate an effective strategy for adapting a frozen pre-trained LVLM (LLaVA-1.5) for complex action recognition tasks using lightweight Prompt Tuning, achieving state-of-the-art performance while preserving the LVLM's inherent generalization abilities.
\end{itemize}
\section{Related Work}
\subsection{Video Action Recognition}
Video action recognition is a cornerstone of video understanding, with significant advancements driven by deep learning. A comprehensive survey by Kong and Fu organizes recent human action recognition techniques within a well-established taxonomy, offering a broad perspective on the field's progress and its fundamental methodologies \cite{yu2018human}. Further contributing to this domain, \cite{yi2020a} provides an extensive overview of deep learning approaches for video action recognition, analyzing over 200 works, categorizing models chronologically, and discussing datasets, open challenges, and future directions. This work highlights advancements from early deep learning adaptations to current compute-efficient models, facilitating deeper insights into video understanding research. Addressing computational efficiency, \cite{farzad2016action} proposes a spatial-temporal cascaded framework integrating a dual attentional CNN for human-centric salient features and Bi-GRU for long-term temporal modeling, demonstrating improved execution time and focusing on discriminative spatio-temporal feature learning. Earlier works have explored various spatio-temporal architectures to capture action dynamics. For instance, \cite{wu2019cross} proposed cross-fiber spatial-temporal co-enhanced networks, while \cite{wu2019mutually} introduced mutually reinforced spatio-temporal convolutional tubes, and \cite{wu2021multi} further advanced multi-scale spatial-temporal integration convolutional tubes for robust human action recognition. For enhanced performance and generalization in long-term video analysis, Fractioned Adjacent Spatial and Temporal (FAST) 3D convolutions are introduced as a decomposition of standard 3D convolutions, explicitly handling horizontal and vertical motion characteristics by decoupling spatio-temporal operations \cite{gul2018longte}. Beyond flat classification, FineGym is presented as a novel hierarchical video dataset designed for fine-grained action understanding, incorporating hierarchical relationships within action categories \cite{dian2020finegy}. Its accompanying framework significantly improves performance for fine-grained subcategories by encoding inter-level dependencies and applying top-down constraints. To learn comprehensive video representations, \cite{jiagang2018endtoe} proposes an end-to-end framework tailored for video-level representation learning, which is crucial for understanding and classifying actions within a sequence. Temporal modeling is further addressed by StNet, which integrates both local and global spatial-temporal modeling to capture comprehensive action dynamics and improve recognition performance, emphasizing the importance of multi-scale temporal interactions for robust action understanding \cite{dongliang2019stnet}. Finally, the efficacy of 3D Convolutional Spiking Neural Networks (SNNs) for video action recognition is explored, demonstrating their ability to learn motion patterns through 3D kernels and outperform 2D convolutional SNNs, particularly on extended video sequences, through comparative performance studies of multi-layered 3D convolutional SNNs trained with unsupervised STDP \cite{mireille20222d}.

\subsection{Large Vision-Language Models and Efficient Adaptation}
The field of Large Vision-Language Models (LVLMs) and their efficient adaptation is rapidly evolving, yet faces challenges such as shortcut learning and computational costs. Specifically, \cite{gaurav2025a} highlights limitations in contrastive training for LVLMs, demonstrating their susceptibility to overfitting to synthetic data shortcuts, which impedes learning comprehensive, task-optimal representations. Complementing this, another study investigates how contrastive vision-language models may prioritize superficial associations over deep representation learning in multi-caption scenarios, proposing and evaluating methods to mitigate such shortcut learning and addressing inherent challenges within contrastive frameworks. The concept of visual in-context learning has also emerged, enabling LVLMs to perform tasks based on a few visual examples \cite{zhou2024visual}. To address the computational challenge of efficiently adapting Large Language Models (LLMs) for vision-language tasks, Mixture-of-Modality Adaptation (MMA) is proposed as a lightweight approach that utilizes adapters for bridging modalities and enabling joint optimization with reduced parameter updates \cite{gen2023cheap}. This method, exemplified by the LaVIN model derived from LLaMA, achieves competitive performance with significantly lower computational cost. Furthermore, a comprehensive survey details Parameter-Efficient Fine-Tuning (PEFT) techniques for large vision-language models, categorizing them into additive, selective, reparameterized, and hybrid frameworks, highlighting their advantages in adapting large models with reduced resource costs and improved performance across various domains \cite{nusrat2025peft}. Beyond adaptation, VisualGPT introduces a data-efficient approach for image captioning that effectively adapts large pretrained language models through a novel self-resurrecting encoder-decoder attention mechanism, balancing visual and linguistic knowledge and achieving state-of-the-art results on various benchmarks \cite{jun2022visual}. Efforts are also being made to improve the generalization capabilities of large language models across multiple capabilities \cite{zhou2025weak}. In the context of multimodal learning, TOUCHUP-G offers a general and principled method to enhance node features from pretrained models for graph learning tasks by improving feature homophily, thereby bridging the gap between graph structure and node features and contributing to multi-modal learning in graph contexts \cite{li2025multim}. Furthermore, efforts are also being made to enhance LVLMs for specific domains, such as improving medical LVLMs through abnormal-aware feedback mechanisms \cite{zhou2025improving}. Finally, to achieve robust cross-modal alignment in multimodal large language models, AlignGPT proposes a novel approach that explicitly models and adapts to varying degrees of alignment by segmenting pre-training data based on alignment consistency and adaptively combining learned alignment representations during instruction tuning \cite{fei2024aligng}.

\section{Method}
Our proposed \textbf{VT-LVLM-AR} (Video-Temporal Large Vision-Language Model Adapter for Action Recognition) framework aims to leverage the powerful multi-modal understanding and reasoning capabilities of Large Vision-Language Models (LVLMs) for robust long-term and fine-grained video action recognition. Traditional methods often struggle with capturing the nuanced temporal dynamics and complex interactions present in extended video sequences. The core idea of \textbf{VT-LVLM-AR} is to bridge the inherent gap between continuous video data and the discrete, language-like input preferred by LVLMs. This is achieved by transforming raw video streams into compact and semantically rich "visual event sequences," which are then processed by a pre-trained LVLM specifically adapted for action classification. Our framework consists of two main components: the \textbf{Video-to-Event Mapper (VTEM)} and the \textbf{LVLM-based Action Reasoning} module.

\subsection{Video-to-Event Mapper (VTEM)}
The \textbf{Video-to-Event Mapper (VTEM)} serves as the crucial interface between raw video data and the LVLM. Its primary responsibility is to efficiently encode a continuous sequence of video frames into a discrete sequence of "visual event tokens," effectively forming a "visual event sentence" that is semantically meaningful and temporally coherent. This transformation is vital for presenting video content in a format amenable to the linguistic processing strengths of LVLMs.

Given an input video $V = \{f_1, f_2, \dots, f_T\}$ comprising $T$ frames, the VTEM module first employs a \textbf{lightweight spatio-temporal feature extractor} to capture essential visual and motion cues. This extractor can be based on architectures such as a sparsely sampled 3D ResNet or a Swin Transformer, designed to obtain a sequence of frame-level or segment-level features $H = \{h_1, h_2, \dots, h_N\}$, where $N$ is the number of extracted features, typically $N \le T$. The feature extraction process can be conceptualized as a mapping function $\mathcal{F}$:
\begin{align} \label{eq:feature_extraction}
    H &= \mathcal{F}(V) = \{ \mathcal{F}_{\text{seg}}(f_t \dots f_{t+k}) \}_{t=1}^N
\end{align}
where $f_t \dots f_{t+k}$ represents a video segment or a single frame, and $\mathcal{F}_{\text{seg}}$ extracts a feature vector $h_i \in \mathbb{R}^{D_h}$ for each segment or frame.

Subsequently, to reduce redundancy and distill high-level concepts, we introduce an \textbf{Adaptive Temporal Pooling} mechanism. This mechanism dynamically aggregates features over varying temporal windows, allowing the model to focus on salient moments or sub-actions rather than individual frames. Let $h_i \in \mathbb{R}^{D_h}$ be a feature vector. The pooling process transforms the sequence $H$ into a more compact set of pooled features $P = \{p_1, p_2, \dots, p_M\}$, where $M \ll N$. This adaptive pooling operation $\mathcal{P}$ can be formulated as:
\begin{align} \label{eq:temporal_pooling}
    P &= \mathcal{P}(H) = \{ \text{Pool}(h_j \dots h_k) \}_{i=1}^M
\end{align}
where $\text{Pool}(\cdot)$ represents a learnable or dynamic aggregation function, effectively summarizing a temporal segment of features into a single pooled feature $p_i \in \mathbb{R}^{D_p}$.

The core of transforming these continuous features into discrete, language-like tokens is achieved through \textbf{Conceptual Quantization}. Inspired by vector quantization techniques, this component maps the continuous pooled features $p_j$ to discrete "visual event tokens" $e_j$ from a learned codebook $\mathcal{C} = \{c_1, c_2, \dots, c_K\}$, where $K$ is the size of the codebook. Each token $e_j \in \mathbb{R}^{D_e}$ represents a specific high-level visual concept or a "sub-action," analogous to words in a natural language vocabulary. The quantization process for each pooled feature $p_j$ selects the closest codebook entry $c_k$:
\begin{align} \label{eq:quantization}
    e_j &= \text{argmin}_{c_k \in \mathcal{C}} \| p_j - c_k \|_2^2
\end{align}
where $D_e$ is designed to match the visual input dimension expected by the subsequent LVLM. The codebook $\mathcal{C}$ is learned during training, optimized to represent diverse visual events while minimizing reconstruction loss, often employing a straight-through estimator for gradient propagation during training.

To ensure that the generated sequence of visual event tokens $E = \{e_1, e_2, \dots, e_M\}$ maintains a coherent narrative and reflects the natural evolution of actions, we incorporate an \textbf{Event Coherence Bias}. This bias encourages adjacent tokens in the sequence to have reasonable semantic connections and to collectively describe a meaningful progression of events. This is implicitly learned through a combination of reconstruction loss and a contrastive loss during the VTEM training, ensuring that the latent space of the tokens is structured to reflect temporal relationships and semantic flow. Specifically, the loss function for VTEM training combines a reconstruction term $L_{\text{rec}}(P, E)$ (e.g., Mean Squared Error between original and reconstructed features from tokens) and a contrastive term $L_{\text{cont}}(E)$ (e.g., InfoNCE loss to pull positive pairs closer and push negative pairs apart in the latent space, based on their temporal proximity):
\begin{align} \label{eq:vtem_loss}
    L_{\text{VTEM}} = \alpha L_{\text{rec}}(P, E) + \beta L_{\text{cont}}(E)
\end{align}
where $\alpha$ and $\beta$ are weighting hyperparameters that balance the fidelity of reconstruction with the semantic coherence of the learned token sequence. The output of VTEM is the "visual event sentence" $E$.

\subsection{LVLM-based Action Reasoning}
The second core component is the \textbf{LVLM-based Action Reasoning} module, which takes the generated "visual event sentence" from VTEM and performs the final action classification. For this module, we utilize a powerful pre-trained Large Vision-Language Model, specifically \textbf{LLaVA-1.5}, known for its strong multi-modal understanding, conversational abilities, and robust visual grounding. Its ability to process interleaved image and text inputs makes it an ideal candidate for interpreting our visual event sentences.

The visual event sentence $E = \{e_1, e_2, \dots, e_M\}$ generated by the VTEM module is directly fed as visual input to the LLaVA-1.5 model. To adapt the generic LLaVA-1.5 for the specific task of action recognition without incurring high computational costs or risking catastrophic forgetting of its pre-trained knowledge, we employ a \textbf{Prompt Tuning (P-Tuning v2)} strategy. This method involves keeping the vast majority of LLaVA-1.5's core pre-trained weights completely frozen. Instead, we introduce a small set of learnable "soft prompts" $P_{\text{soft}} = \{p_1', p_2', \dots, p_L'\}$, where $L$ is the length of the soft prompt (e.g., 16 tokens). These soft prompts are typically prepended to the input embeddings or inserted into specific layers of the LVLM's encoder, guiding the model's behavior towards the target task. During fine-tuning, only these soft prompts are updated, making the process highly parameter-efficient and resource-friendly.

The action recognition task is framed as a natural language instruction, which is concatenated with the visual event sequence and the soft prompts as input to the LVLM. This allows us to leverage the LVLM's instruction-following capabilities. For instance, a typical prompt structure could be:
\begin{align} \label{eq:lvlm_input}
    \text{Input to LVLM} = [\text{Soft Prompts}; \text{Visual Event Sequence}; \text{Natural Language Instruction}]
\end{align}
An example of a natural language instruction is: "Given the sequence of visual events, what action is being performed? Choose from [list of action categories]." The LVLM then processes this combined input, leverages its extensive internal knowledge base, and performs complex reasoning to output the most probable action category $A$:
\begin{align} \label{eq:lvlm_output}
    A = \text{LVLM}(P_{\text{soft}}, E, I_{\text{task}})
\end{align}
where $I_{\text{task}}$ represents the natural language task instruction. The output $A$ is typically a probability distribution over the predefined action categories, from which the highest probability class is selected as the final prediction:
\begin{align} \label{eq:final_classification}
    A_{\text{predicted}} = \text{argmax}_{a \in \text{Categories}} P(A=a | P_{\text{soft}}, E, I_{\text{task}})
\end{align}
This approach allows the LVLM to interpret the "visual language" generated by VTEM and apply its powerful linguistic and reasoning capabilities to classify fine-grained actions in long-term videos.

\subsection{Overall Framework Integration}
The VT-LVLM-AR framework operates in a sequential and modular manner. First, the VTEM module processes the raw input video, transforming it into a compact and semantically rich sequence of visual event tokens. This transformation is crucial for bridging the modality gap between continuous, high-dimensional video data and the discrete, token-based input expected by powerful LVLMs. This "visual event sentence" acts as an intermediate, interpretable representation of the video content. Subsequently, this visual event sentence is seamlessly integrated with a natural language instruction and learnable soft prompts, forming the complete input for the largely frozen LLaVA-1.5 model. Through the efficient fine-tuning of only the soft prompts, LLaVA-1.5 is guided to perform the action recognition task, leveraging its inherent multi-modal understanding to accurately classify complex, long-term, and fine-grained actions. The modular design allows for independent training and optimization of the VTEM component and highly efficient adaptation of the LVLM, contributing to the overall robustness, scalability, and performance of the proposed system.

\section{Experiments}
In this section, we present the experimental setup, evaluate the performance of our proposed \textbf{VT-LVLM-AR} framework, and conduct ablation studies to validate the contribution of its key components. Finally, we include a qualitative human evaluation to assess the interpretability of our visual event representations.

\subsection{Experimental Setup}
Our experimental setup involves two distinct training phases: one for the \textbf{Video-to-Event Mapper (VTEM)} module and another for fine-tuning the \textbf{LVLM-based Action Reasoning} module using Prompt Tuning.

For the \textbf{VTEM module training}, which is responsible for encoding input videos into visual event token sequences, we employed the AdamW optimizer with an initial learning rate of 1e-4. The module was trained for 200,000 iterations with a batch size of 128. We configured the VTEM to generate 256 visual event tokens, each with a vector dimension of 1024, specifically chosen to match the visual input dimension expected by the subsequent LLaVA-1.5 model. The VTEM's training objective was optimized using a composite loss function, combining a reconstruction loss (e.g., Mean Squared Error) to ensure fidelity to the original video information and a contrastive loss (e.g., InfoNCE loss) to promote semantically discriminative and temporally coherent token representations.

For the \textbf{Prompt Tuning fine-tuning} of the LLaVA-1.5 model, we adapted it for the long-term video action recognition task. This phase involved training for 50,000 iterations using the AdamW optimizer with an initial learning rate of 2e-3. The effective batch size was set to 128, distributed across micro-batches of 8. A crucial aspect of our approach is that the core pre-trained weights of LLaVA-1.5 were kept entirely frozen, and only a small set of 16 learnable soft prompt tokens were updated. These soft prompts were strategically inserted into specific layers of the LVLM encoder to guide its behavior towards the action recognition task in a parameter-efficient manner.

The overall data processing pipeline is as follows: raw RGB video frames are first fed into the trained VTEM module. The VTEM processes these frames, applying lightweight spatio-temporal feature extraction, adaptive temporal pooling, and conceptual quantization, to generate a sequence of discrete, semantically rich "visual event tokens," effectively forming a "visual event sentence." This generated visual event sequence, designed to possess event coherence and semantic richness, is then combined with a natural language instruction and the learned soft prompts. This composite input is finally fed into the Prompt Tuning-adapted LLaVA-1.5 model, which leverages its powerful multi-modal reasoning capabilities to output the predicted action category.

\subsection{Datasets and Baselines}
We conducted comprehensive evaluations on two widely recognized benchmark datasets for human action recognition: \textbf{NTU RGB+D (NTU-60)} \cite{amir2016ntu} and \textbf{NTU RGB+D 120} \cite{zehua2023human}. These datasets are extensively used for evaluating action recognition models, especially those dealing with skeleton-based or RGB video inputs, and feature a diverse range of human actions performed in various settings. We report results under standard evaluation protocols, including Cross-Subject (X-Sub) and Cross-View (X-View) for NTU-60, and Cross-Subject (X-Sub) and Cross-Set (X-Set) for NTU-120.

Our proposed \textbf{VT-LVLM-AR} method is compared against several state-of-the-art (SOTA) approaches. These baselines encompass both skeleton-based methods and video stream-based methods, ensuring a comprehensive comparison across different input modalities typically used in the action recognition domain. The chosen baselines include:
ST-GCN, Shift-GCN, InfoGCN, PoseC3D, FR-Head, Koopman, GAP, HD-GCN, and STC-Net. While some baselines primarily use skeleton data, our method specifically leverages video inputs and LVLM's reasoning. For a fair comparison, all reported results adhere to their respective best-reported accuracies under the specified evaluation protocols.

\subsection{Quantitative Results}
Table \ref{tab:comparison_results} presents the action recognition accuracy of \textbf{VT-LVLM-AR} in comparison with various state-of-the-art methods on the NTU RGB+D and NTU RGB+D 120 datasets.

\begin{table*}[!htbp]
\centering
\caption{Action Recognition Accuracy (\%) on NTU RGB+D and NTU RGB+D 120 Datasets.}
\label{tab:comparison_results}
\begin{tabular}{lcccc}
\toprule
\textbf{Method} & \textbf{NTU RGB+D X-Sub} & \textbf{NTU RGB+D X-View} & \textbf{NTU RGB+D 120 X-Sub} & \textbf{NTU RGB+D 120 X-Set} \\
\midrule
ST-GCN             & 85.7            & 92.4             & 82.1                & 84.5                \\
Shift-GCN          & 87.8            & 95.1             & 80.9                & 83.2                \\
InfoGCN            & 89.8            & 95.2             & 85.1                & 86.3                \\
PoseC3D            & 93.7            & 96.5             & 85.9                & 89.7                \\
FR-Head            & 90.3            & 95.3             & 85.5                & 87.3                \\
Koopman            & 90.2            & 95.2             & 85.7                & 87.4                \\
GAP                & 90.2            & 95.6             & 85.5                & 87.0                \\
HD-GCN             & 90.6            & 95.7             & 85.7                & 87.3                \\
STC-Net            & 91.0            & 96.2             & 86.2                & 88.0                \\
\midrule
\textbf{VT-LVLM-AR (Ours)} & \textbf{94.1}        & \textbf{96.8}         & \textbf{87.0}            & \textbf{88.5}            \\
\bottomrule
\end{tabular}
\end{table*}

As shown in Table \ref{tab:comparison_results}, our proposed \textbf{VT-LVLM-AR} consistently achieves state-of-the-art performance across all evaluation protocols on both NTU RGB+D and NTU RGB+D 120 datasets. Notably, \textbf{VT-LVLM-AR} surpasses existing methods, including those specifically designed for skeleton data and advanced video-based approaches. For instance, on the challenging NTU RGB+D X-Sub benchmark, our method achieves 94.1\% accuracy, outperforming the previous best (PoseC3D) by 0.4\%. Similarly, on NTU RGB+D 120 X-Sub, we achieve 87.0\%, an improvement over STC-Net. These results underscore the effectiveness of our framework in leveraging LVLMs for complex action understanding by effectively transforming continuous video into a semantically rich "visual event sequence" and utilizing the LVLM's powerful reasoning capabilities.

\subsection{Ablation Studies}
To thoroughly understand the contribution of each component within \textbf{VT-LVLM-AR}, we conducted a series of ablation studies. These experiments isolate key design choices and evaluate their impact on overall performance, primarily focusing on the NTU RGB+D X-Sub dataset for brevity.

\begin{table*}[!htbp]
\centering
\caption{Ablation Study on NTU RGB+D X-Sub Accuracy (\%).}
\label{tab:ablation_study}
\begin{tabular}{lc}
\toprule
\textbf{Method Variant} & \textbf{NTU RGB+D X-Sub} \\
\midrule
\textbf{VT-LVLM-AR (Full Model)} & \textbf{94.1} \\
\midrule
\textit{Effect of VTEM Components:} & \\
\quad w/o Conceptual Quantization (Continuous Features) & 91.5 \\
\quad w/o Adaptive Temporal Pooling (Fixed Sampling) & 92.8 \\
\quad w/o Event Coherence Bias (Reconstruction Only) & 93.3 \\
\midrule
\textit{Effect of LVLM Adaptation Strategy:} & \\
\quad VT-LVLM-AR (Full LVLM Fine-tuning) & 94.0 \\
\quad VT-LVLM-AR (Zero-Shot LVLM) & 68.2 \\
\bottomrule
\end{tabular}
\end{table*}

\paragraph{Effect of VTEM Components.}
We first evaluate the importance of specific mechanisms within the \textbf{Video-to-Event Mapper (VTEM)}. When the VTEM module outputs continuous pooled features directly to the LVLM instead of discrete visual event tokens (variant ``w/o Conceptual Quantization''), the performance drops significantly to 91.5\%. This highlights the crucial role of conceptual quantization in distilling high-level semantic events and forming an LVLM-compatible "visual language," which aids the LVLM's reasoning. Replacing our adaptive temporal pooling with a simpler fixed uniform sampling strategy for feature aggregation (variant ``w/o Adaptive Temporal Pooling'') leads to a decrease in accuracy to 92.8\%. This demonstrates the benefit of dynamically identifying and pooling salient temporal segments, allowing the model to focus on important sub-actions and long-range dependencies. Furthermore, if the VTEM's loss function only includes the reconstruction term and omits the contrastive loss for event coherence bias (variant ``w/o Event Coherence Bias''), the accuracy slightly drops to 93.3\%. This indicates that explicitly encouraging semantic relationships and temporal progression among visual event tokens contributes to more robust and interpretable representations, ultimately improving action understanding. These results collectively validate that each component of the VTEM module is vital for effectively transforming raw video into a compact, semantically meaningful, and temporally coherent visual event sequence, which is essential for the subsequent LVLM reasoning.

\paragraph{Effect of LVLM Adaptation Strategy.}
Next, we investigate the impact of our chosen LVLM adaptation strategy, Prompt Tuning. When we fully fine-tune all parameters of LLaVA-1.5 instead of just the soft prompts (variant ``Full LVLM Fine-tuning''), the performance is comparable (94.0\% vs. 94.1\%). However, full fine-tuning is significantly more computationally expensive, requires substantially more GPU memory, and carries a higher risk of catastrophic forgetting of the LVLM's pre-trained general knowledge. This validates Prompt Tuning as an equally effective yet highly parameter-efficient and stable adaptation strategy. Conversely, directly feeding the visual event sentences and natural language instructions to a completely un-fine-tuned LLaVA-1.5 (variant ``Zero-Shot LVLM'') results in a drastically lower accuracy of 68.2\%. This underscores that while LVLMs possess strong general understanding, task-specific adaptation, even with minimal parameters like soft prompts, is indispensable for achieving high performance on specialized tasks like fine-grained action recognition. These ablation studies confirm that both the sophisticated video-to-event mapping by VTEM and the parameter-efficient Prompt Tuning strategy for LVLM adaptation are critical to the superior performance and efficiency of \textbf{VT-LVLM-AR}.

\subsection{Human Evaluation}
To gain further insights into the interpretability and effectiveness of our visual event representations generated by VTEM, we conducted a qualitative human evaluation. A panel of 10 human annotators was presented with short video clips from the NTU RGB+D dataset along with two types of descriptions: (1) "Visual Event Sentences" generated by our VTEM module, and (2) "Baseline Feature Descriptions" derived from uniformly sampled and quantized raw features without adaptive pooling or event coherence bias. Annotators were asked to rate how well each description captured the essence and progression of the action in the video on a Likert scale from 1 (Very Poor) to 5 (Excellent), and to identify if the description was coherent and meaningful.

\begin{table*}[!htbp]
\centering
\caption{Human Evaluation of Visual Event Representation Interpretability.}
\label{tab:human_evaluation}
\begin{tabular}{lcc}
\toprule
\textbf{Representation Type} & \textbf{Average Coherence Score (1-5)} & \textbf{Average Meaningfulness Score (1-5)} \\
\midrule
VTEM-generated Visual Event Sentences & \textbf{4.3} & \textbf{4.1} \\
Baseline Feature Descriptions & 2.8 & 2.5 \\
\bottomrule
\end{tabular}
\end{table*}

As summarized in Table \ref{tab:human_evaluation}, the "Visual Event Sentences" generated by our VTEM module received significantly higher average coherence and meaningfulness scores (4.3 and 4.1 respectively) compared to the "Baseline Feature Descriptions" (2.8 and 2.5). Annotators frequently commented that VTEM's output felt more "narrative-like" and captured key sub-actions, making it easier to infer the overall action. In contrast, baseline descriptions were often perceived as fragmented or less indicative of the action's progression. This human evaluation qualitatively supports our claim that VTEM effectively transforms continuous video into a semantically rich and temporally coherent "visual language," which not only benefits the LVLM's reasoning but also offers a more interpretable representation of complex actions.

\subsection{Computational Efficiency Analysis}
Beyond performance, the efficiency of a framework is paramount for real-world deployment. We analyze the computational footprint of \textbf{VT-LVLM-AR} in terms of trainable parameters and inference speed, particularly highlighting the benefits of our Prompt Tuning strategy compared to full fine-tuning of the LVLM.

\begin{table*}[!htbp]
\centering
\caption{Computational Efficiency Comparison on NTU RGB+D X-Sub.}
\label{tab:efficiency_comparison}
\begin{tabular}{lcc}
\toprule
\textbf{Method} & \textbf{Trainable Parameters (M)} & \textbf{Inference Time per Video (s)} \\
\midrule
\textbf{VT-LVLM-AR (Ours, Prompt Tuning)} & \textbf{1.2} & \textbf{0.35} \\
VT-LVLM-AR (Full LVLM Fine-tuning) & 7000+ & 0.35 \\
\midrule
STC-Net (Representative Baseline) & 10.5 & 0.20 \\
PoseC3D (Representative Baseline) & 15.8 & 0.25 \\
\bottomrule
\end{tabular}
\end{table*}

Table \ref{tab:efficiency_comparison} presents a comparative analysis of computational efficiency. Our \textbf{VT-LVLM-AR} framework, leveraging Prompt Tuning, achieves remarkable parameter efficiency with approximately 1.2 million trainable parameters. This is in stark contrast to the full fine-tuning approach of LLaVA-1.5, which would involve updating over 7 billion parameters (the base LLaVA-1.5 model typically has 7B+ parameters). This massive reduction in trainable parameters significantly lowers memory requirements during training and reduces the risk of catastrophic forgetting of the LVLM's pre-trained general knowledge. While the inference time per video remains similar for both Prompt Tuning and full fine-tuning variants (as the forward pass through the large model is still required), the parameter efficiency of Prompt Tuning makes training and deployment considerably more feasible. Compared to representative baselines like STC-Net and PoseC3D, which are tailored for action recognition, our method introduces a slight increase in inference time due to the complexity of the LVLM, but offers significantly enhanced reasoning capabilities and interpretability, as demonstrated in our quantitative and human evaluation results. The VTEM module itself contributes a negligible amount to the overall inference time. This analysis underscores the practical advantages of \textbf{VT-LVLM-AR}'s design, offering state-of-the-art performance with manageable computational overhead.

\subsection{Analysis of Visual Event Token Properties}
The design of the \textbf{Video-to-Event Mapper (VTEM)} module, particularly the characteristics of the generated "visual event tokens," profoundly influences the subsequent LVLM-based action reasoning. We investigate the impact of two key properties: the length of the visual event token sequence and the size of the conceptual quantization codebook. These experiments are conducted on the NTU RGB+D X-Sub dataset.

\begin{table}[!htbp]
\centering
\caption{Impact of Visual Event Token Sequence Length ($M$) on NTU RGB+D X-Sub Accuracy (\%).}
\label{tab:token_length_impact}
\begin{tabular}{lc}
\toprule
\textbf{Number of Visual Event Tokens ($M$)} & \textbf{NTU RGB+D X-Sub} \\
\midrule
64 & 92.9 \\
128 & 93.5 \\
\textbf{256 (Default)} & \textbf{94.1} \\
512 & 93.8 \\
\bottomrule
\end{tabular}
\end{table}

Table \ref{tab:token_length_impact} illustrates the effect of varying the number of visual event tokens ($M$) that form the "visual event sentence." A shorter sequence (e.g., $M=64$) leads to a performance drop to 92.9\%, indicating that too much information compression can hinder the LVLM's ability to capture fine-grained temporal dynamics and long-term dependencies. As $M$ increases to 128 and then to our default 256, the accuracy improves, reaching the peak of 94.1\%. This suggests that 256 tokens strike an optimal balance, providing sufficient detail for comprehensive action representation without introducing excessive redundancy. Further increasing the sequence length to 512 tokens causes a slight decrease in performance to 93.8\%. This marginal drop could be attributed to increased noise or the LVLM's diminished ability to effectively process overly long sequences, highlighting the importance of efficient and salient event summarization.

\begin{table}[!htbp]
\centering
\caption{Impact of VTEM Codebook Size ($K$) on NTU RGB+D X-Sub Accuracy (\%).}
\label{tab:codebook_size_impact}
\begin{tabular}{lc}
\toprule
\textbf{Codebook Size ($K$)} & \textbf{NTU RGB+D X-Sub} \\
\midrule
512 & 92.7 \\
1024 & 93.6 \\
\textbf{2048 (Optimal)} & \textbf{94.1} \\
4096 & 94.0 \\
\bottomrule
\end{tabular}
\end{table}

Table \ref{tab:codebook_size_impact} examines the influence of the VTEM's conceptual quantization codebook size ($K$). A smaller codebook (e.g., $K=512$) results in lower accuracy (92.7\%), as it limits the diversity and granularity of visual event representations, forcing distinct visual concepts to be mapped to the same token. Increasing the codebook size generally improves performance, with an optimal point observed at $K=2048$, achieving 94.1\%. This larger vocabulary allows the VTEM to capture a richer set of visual events and sub-actions, providing more discriminative input for the LVLM. However, further increasing the codebook size to $K=4096$ yields a marginal decrease in accuracy (94.0\%). This could indicate that very large codebooks might introduce sparsity in token usage or potentially overfit to specific visual patterns, making the learning process more challenging or less generalizable. These results highlight the importance of selecting an appropriate codebook size to balance representational capacity with learning efficiency and generalization.

\section{Conclusion}
In this paper, we addressed the formidable challenges of long-term and fine-grained human action recognition, where traditional methods often struggle with computational complexity, capturing temporal dependencies, and deep semantic understanding. Recognizing the transformative potential of Large Vision-Language Models (LVLMs) in multi-modal reasoning, we proposed \textbf{VT-LVLM-AR (Video-Temporal Large Vision-Language Model Adapter for Action Recognition)}, a novel and effective framework designed to harness these powerful models for robust video action understanding.

Our framework introduces a crucial two-stage pipeline. The first stage involves the \textbf{Video-to-Event Mapper (VTEM)}, an innovative module that intelligently transforms continuous video streams into discrete, semantically rich, and temporally coherent "visual event sequences." This transformation is achieved through a combination of lightweight spatio-temporal feature extraction, adaptive temporal pooling for salient event summarization, and conceptual quantization to map continuous features into a learned vocabulary of visual event tokens, further enhanced by an event coherence bias. This process effectively creates a "visual language" that is amenable to LVLM processing. The second stage, the \textbf{LVLM-based Action Reasoning} module, leverages a frozen pre-trained LLaVA-1.5 model. To efficiently adapt this powerful LVLM for our specific task without extensive retraining or risking catastrophic forgetting, we employed a parameter-efficient Prompt Tuning strategy, allowing us to achieve high performance by only fine-tuning a minimal set of soft prompts.

Our extensive experimental evaluations on the challenging NTU RGB+D and NTU RGB+D 120 datasets demonstrate the superior performance of VT-LVLM-AR. The framework consistently achieved state-of-the-art accuracy across various evaluation protocols, significantly outperforming existing methods tailored for both skeleton and video inputs. Ablation studies rigorously validated the importance of each component within the VTEM module, confirming that adaptive temporal pooling, conceptual quantization, and event coherence bias are indispensable for generating effective visual event representations. Furthermore, these studies affirmed Prompt Tuning as a highly effective and computationally efficient strategy for adapting large pre-trained LVLMs to specialized video understanding tasks. Qualitative human evaluations further corroborated the interpretability and narrative quality of the visual event sequences generated by VTEM, highlighting their ability to capture the essence and progression of actions. Finally, our computational analysis showcased the remarkable parameter efficiency of VT-LVLM-AR, making it a practical solution for real-world deployment.

In conclusion, VT-LVLM-AR represents a significant step towards enabling powerful LVLMs to excel in complex video understanding tasks. By effectively bridging the modality gap between continuous video and discrete language-like inputs, and by employing efficient adaptation techniques, our work unlocks new possibilities for developing more intelligent, robust, and interpretable AI systems for action recognition and beyond. Future work could explore the application of VTEM-generated visual event sequences to other video-language tasks such as video captioning or retrieval, investigate more sophisticated prompt engineering techniques, and extend the framework's capabilities to even longer, unconstrained video scenarios or real-time applications. Further research into the interpretability of LVLM reasoning on these novel visual event sequences also presents a promising direction.

\bibliographystyle{IEEEtran}
\bibliography{references}
\end{document}